# Automatic Techniques for Gridding cDNA Microarray Images

Naima Kaabouch[1], *Member, IEEE*, and Hamid Shahbazkia
[1]Department of Electrical Engineering, University of North Dakota
Grand Forks, ND 58202-7165
[2]University of Algarve, FCT, Campus de Gambelas
8000 Faro, Portugal

*Abstract*—Microarray is considered an important instrument and powerful new technology for large-scale gene sequence and gene expression analysis. One of the major challenges of this technique is the image processing phase. The accuracy of this phase has an important impact on the accuracy and effectiveness of the subsequent gene expression and identification analysis. The processing can be organized mainly into four steps: gridding, spot isolation, segmentation, and quantification. Although several commercial software packages are now available, microarray image analysis still requires some intervention by the user, and thus a certain level of image processing expertise. This paper describes and compares four techniques that perform automatic gridding and spot isolation. The proposed techniques are based on template matching technique, standard deviation, sum, and derivative of these profiles. Experimental results show that the accuracy of the derivative of the sum profile is highly accurate compared to other techniques for good and poor quality microarray images.

## I. INTRODUCTION

Microarray is considered an important instrument and powerful new technology for large-scale gene sequence and gene expression analysis. Originally, the technique evolved from E. Southern's technique in the 1970s [1] and became modernized in the last decade by two key innovations. One was pioneered by P. Brown [2], consisting of the use of nonporous solid support to facilitate miniaturization and fluorescent-based detection. The second was the development of methods for high-density spatial synthesis of oligonucleotides [3]. The resulting microarray images contain thousands of genes that can be used to survey the DNA and RNA variations [4], which will someday become a standard tool for both molecular biology research and genomic clinical diagnosis, such as cancer diagnosis [5, 6] and diabetes diagnosis [7, 8].

One of the major challenges of the microarray technique is the image processing phase. The purpose of this phase is to extract each spotted DNA sequence as well as to obtain background estimates and quality measures. The accuracy of this phase has an important impact on the accuracy and effectiveness of the subsequent gene expression and identification analysis. The processing in this phase can be organized mainly into three steps: 1) gridding and extraction of individual spots; 2) segmentation of the spot images; 3) quantification. The performance of these steps is critical, since this process will directly impact the strategy and quality of downstream microarray data. Because of the noise that affects microarray images, the image processing is analytically complex and labor intensive. Thus, it is highly important that these steps are automated, in order to save time and resources and especially to remove user-dependent variations.

Commercial systems that perform some of these steps – such as ScanAlyze, GenePix, and QuantArray– already exist. However, most of these software systems are semiautomatic. For example, for gridding they require the end-users to specify the geometry of the array, such as number of rows and columns. This manual setting works for very good images, but for average and poor quality images, the manual gridding is time consuming and rarely correctly done, which can affect not only the spot extraction but also its grayscale intensity and, hence, the quality of gene expression data.

A few research papers have been published to address automatic gridding. Jain *et al.* [9] used a gridding algorithm based on axis projections of image intensity. Yang *et al.* [10] used template matching and seeded region growing methods for gridding. Other authors proposed morphological methods for grid segmentation [11]. Since these approaches use templates or employ axis projections as a central component, irregular and overlapping grid layouts may cause problems.

In this paper we describe and compare different techniques to address the problem of automatic gridding and spot extraction. These techniques are based on a template matching, the horizontal and vertical sums and standard deviations, as well as the derivatives of these profiles.

II. METHODOLOGY

Microarray images are grouped in subarrays, with each subarray containing several hundred spots. Thus, to extract each spot, the image has to be segmented first in subarrays.. We implemented and compared several techniques to divide the image into subarrays and to extract each spot

- Template matching
- Horizontal and vertical sum profiles
- Horizontal and vertical standard deviation profiles
- Derivatives of the horizontal and vertical sum profiles or of the standard deviation profiles

A. *Template matching*

This technique uses a predefined image template to grid the subarray data image, thereby isolating each spot.

B. *Horizontal and vertical sum profiles*

Horizontal and vertical sum signals, by indicating the maxima and minima, point to the locations of subarrays and the gaps between them, respectively (Figs. 3 and 4). Thus, the gridding can be done by locating the middle of the gaps between subarrays. Then to extract spots, the same aforementioned technique is applied to each subarray image with the maxima representing the spots and the minima the gaps between them.

The horizontal sum profile is obtained by adding the pixel intensities of every row. The sum profile is defined as follows:

$$S(i) = \sum_{j=1}^{N} I_{ij} \qquad (1)$$

Where, $i$ and $j$ are the indices of the lane columns and rows, respectively.

C. *Horizontal and vertical standard deviation*

This technique uses the same methodology as in section B., but applies it to the standard deviation profiles. These profiles are also used to locate the middle of the gaps between subarray and between spots. The horizontal profile is obtained by computing the standard deviation of pixel intensities for every column. Horizontal standard deviation is defined as follows:

$$\sigma(i) = \frac{1}{M-1} \sum_{j=1}^{N} (I_{ij} - \overline{I_i})^2)^{1/2} \qquad (2)$$

Where, $N$ and $M$ are the image numbers of rows and columns, respectively, $i$ and $j$ are the indices of the image columns and rows, respectively. $I_{ij}$ is the intensity of the pixel located at column $i$ and row $j$. $\overline{I_i}$ represents the mean of the intensity values of row $j$ and is defined follows:

$$\overline{I_i} = \frac{1}{N} \sum_{j=1}^{N} I_{ij} \qquad (3)$$

D. *First derivatives of the standard deviation or the sum profiles*

Because the subarrays are uniformly distributed, as are the spots in each subarray, the horizontal and vertical signals of any profile are periodic. However, because of the noise, the backgrounds of the microarray data images are not uniform and, thus, the gridding requires setting a threshold depending on the quality of the image. This threshold can be avoided by computing the derivative of these profiles. The numerical derivative profile is computed using the following equation:

$$P'(i) = \frac{\partial}{\partial i} P(i) = P(i+1) - P(i) \qquad (4)$$

Where, $i$ and $j$ are the indices of the image columns and rows, respectively. The periods of the horizontal and vertical profiles give the width and the height of the spot images, respectively.

Although this technique can be used to grid the image into subarrays, the algorithm was applied only to extract spots from subarrays. The algorithm is as follows

- Compute the first and second derivative of the vertical profile
- Locate the pixels $(x_1, x_2, ... x_n)$ where the first derivative is zero and the second derivative is positive (see Fig. 2)
- Use these coordinates to divide the subarray vertically
- Compute the first and second derivatives of the vertical profile
- Locate the pixels where the first derivative is zero and the second derivative is positive
- Use these coordinates to divide the subarray horizontally

Using techniques B, C, and D, the gridding is done by detecting the middle of the gaps between signals corresponding to the sub-arrays both horizontally and vertically. Once the sub-arrays are isolated, the same technique can be applied horizontally and vertically to each subarray in order to isolate the spots. The horizontal and vertical periods of any signal correspond to the width and height of each spot, respectively.

## III. RESULTS

We have implemented several techniques to grid cDNA microarray images and compared their performances using a set microarray images of different qualities.

Fig. 1 shows a typical cDNA microarray image obtained from yeast and containing 16 subarrays. Each subarray contains 384 spots, with a total of 6,144 spots that have to be isolated automatically. This image is an example of an average quality image, where the brightness of the spots is highly non-uniform. Most of the spots have intensity levels close to background levels, which make the gridding and spot extraction a difficult and tedious task.

Figs. 4 a) and 4 b) show the horizontal and vertical profiles of the sum and standard deviation, respectively, for the microarray data image of Fig. 3. These profiles are binarized by setting a threshold level depending on the quality of the image. The resulting profile is then used to determine the middle of the gaps between signals, as shown in Fig. 4. The results of gridding show high efficiency in segmenting the image in subarrays; however, the techniques based on the horizontal sum and standard deviation profiles are not fully automatic and can fail when used on average and poor quality images. This failure can occur because the gaps between signals corresponding to adjacent spots are very small and can be easily affected by the noise, as shown in Fig. 5 and Fig. 6. Additionally, end users are required to set a threshold level for the noise and, hence, need to possess a certain expertise in image processing. The application of the template matching technique also gives good results but fails for images that present geometric distortion.

The setting of the threshold level is avoided by computing the derivative of the horizontal sum and standard deviation profiles. As shown in Fig. 7 and Fig. 8, corresponding to the derivatives of the horizontal sum and standard deviation profiles, no threshold is needed. In each interval period, the maximum and minimum of the signal correspond to the left and right edges of the spot. The period of the signals in Fig. 7 and Fig. 8 gives the width of the spot image, while the period of the vertical signal gives the height of the spot image. Figs. 9 and 10 give the result of the gridding corresponding to subarray number 1 on the top left of Fig.3, using the sum profile (or standard deviation profile) and the derivative of the sum profile (or derivative of the standard deviation profile), respectively. Comparing the two last rows in the two images, one can conclude that the two techniques B and D fail to grid correctly because several cells in this last row of Fig. 9 contain more than one spot while the cells in the last row of Fig. 10, using the derivative profile, contain only one spot per cell. This technique, derivative, showed better results than the other techniques, both in gridding and extracting individual spots. However, the derivative of the sum profile gives higher signals and, thus, better results in gridding and extracting spots.

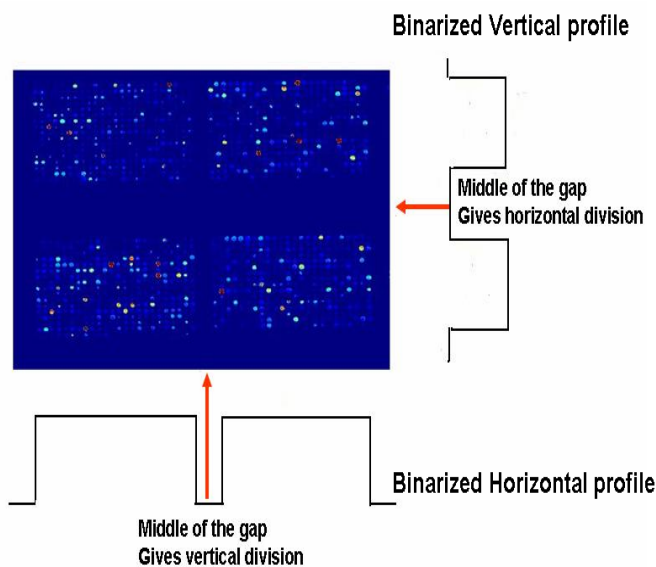

Fig. 1: Methodoly for gridding an image consisting of four subarrays.

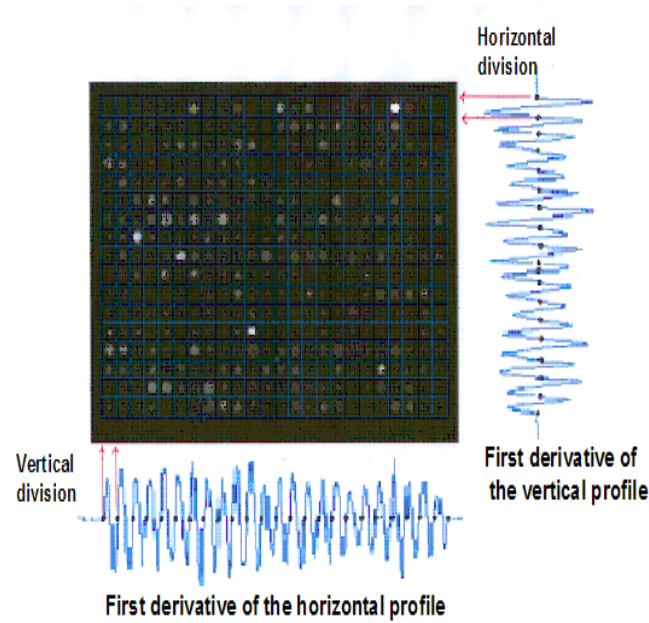

Fig.2: Methodology of gridding a subarray using the first derivative of sum or standard deviation profiles.

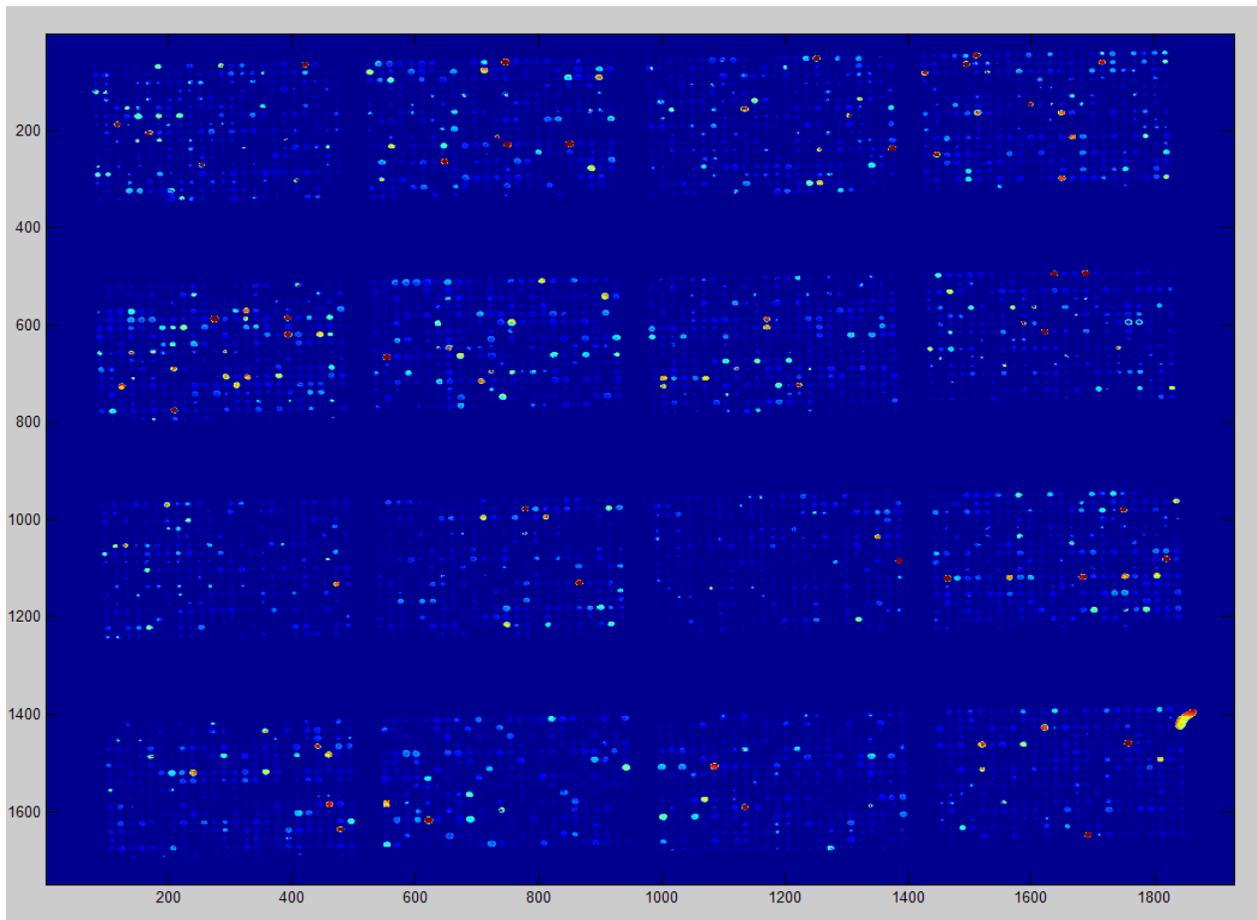

Fig. 3. Example of microarray image containing 16 subarrays.

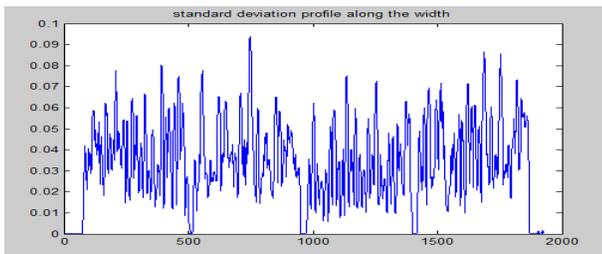
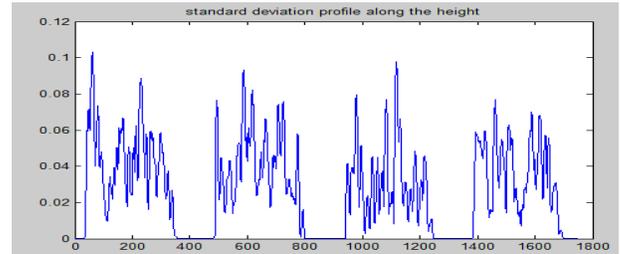

Fig. 4. Typical profile of a) the horizontal and b) the vertical standard deviation and the sum profiles of Fig. 3.

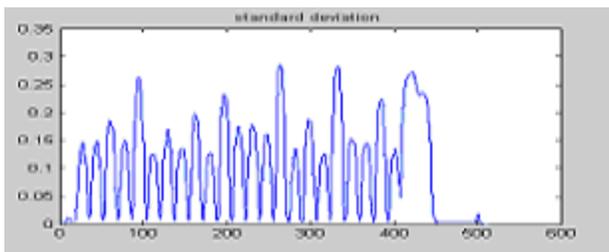
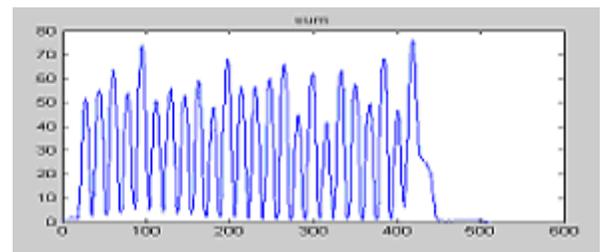

Fig. 5. Profile of the horizontal standard deviation

Fig. 6. Profile of the horizontal sum profile.

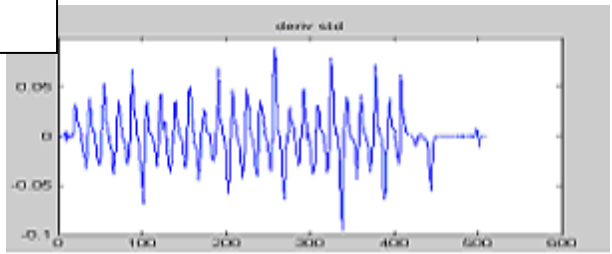

Fig. 7. Derivative of the horizontal standard deviation profile.

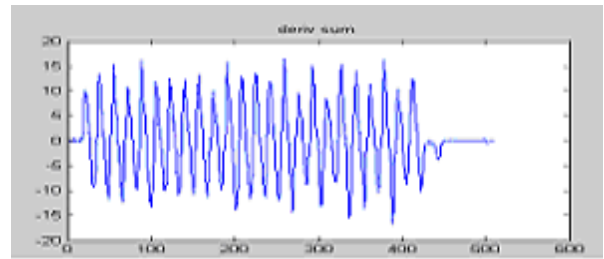

Fig. 8. Derivative of the horizontal sum profile.

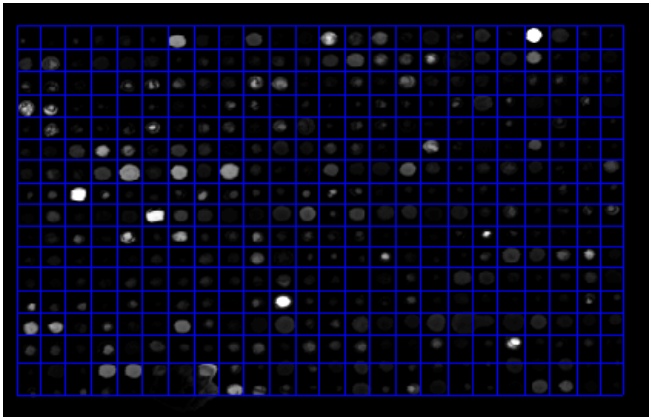

Fig. 9. Gridded subarray number one in Fig. 3 using the horizontal and vertical sum (or standard deviation) profiles.

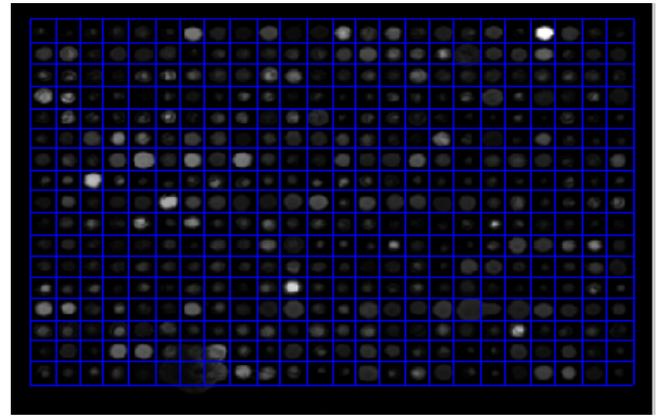

Fig. 10. Gridded subarray number one in Fig. 3 using the derivative of the horizontal and vertical sum (or standard deviation) profiles.

CONCLUSION

In this work, we have presented a new technique that automatically subdivides the microarray images in subarrays and isolates the spots in each subarray image. We have compared the robustness and performance of this technique with other existing algorithms using a set of cDNA microarray images of different qualities and with different spot shapes. This evaluation and comparison showed that the derivative of the sum of profile technique is the most accurate for gridding all types of microarray data images. The proposed technique is not affected by the noise or the shape of the spots, and does not require any images. The proposed technique is not affected by the noise or the shape of the spots, and does not require any threshold level. However, if the image or the subarrays are misaligned an alignment algorithm is required before gridding the image or the subarrays. Our objective in the future is to address this issue and to develop criteria to evaluate the proposed algorithms.